\documentclass{article}

\usepackage[nonatbib]{nips_2018}

\usepackage[utf8]{inputenc} 
\usepackage[T1]{fontenc}    
\usepackage{hyperref}       
\usepackage{url}            
\usepackage{booktabs}       
\usepackage{amsfonts}       
\usepackage{nicefrac}       
\usepackage{microtype}      
\usepackage{authblk}
\usepackage{chngcntr}       
\usepackage{bm}             
\usepackage[pdftex]{graphicx}
\usepackage{color}
\usepackage{footnote}

\begin{document}

\title{Capsule Networks with Max-Min Normalization}
\maketitle

\begin{abstract}
	Capsule Networks (CapsNet) use the Softmax function to convert the logits of the routing coefficients into a set of normalized values that signify the assignment probabilities between capsules in adjacent layers. We show that the use of Softmax prevents capsule layers from forming optimal couplings between lower and higher-level capsules. Softmax constrains the dynamic range of the routing coefficients and leads to probabilities that remain mostly uniform after several routing iterations. Instead, we propose the use of Max-Min normalization. Max-Min performs a scale-invariant normalization of the logits that allows each lower-level capsule to take on an independent value, constrained only by the bounds of normalization. Max-Min provides consistent improvement in test accuracy across five datasets and allows more routing iterations without a decrease in network performance. A single CapsNet trained using Max-Min achieves an improved test error of $0.20\%$ on the MNIST dataset. With a simple 3-model majority vote, we achieve a test error of $0.17\%$ on MNIST.
\end{abstract}

\section{Introduction}
\label{Section: Introduction}
Deep learning systems are powerful tools for recognition, prediction and strategy in fields such as vision, speech, language and games \cite{Lecun_2015, Silver_2017}. The mammalian visual system extracts features from objects in cluttered scenes and then combines them for robust recognition. Inversion of sensory processing, using the ubiquitous feedback pathways present throughout sensory systems, is a major component of object recognition \cite{Harth_1987}. Learning systems that utilize compositionality of objects \cite{Lake_2015}, along with dynamic binding of parts (or features) to wholes (or objects) can become power architectures. Capsule Networks (CapsNets) have the potential to perform object recognition in a natural and systematic fashion.

Capsule Networks \cite{Sabour_2017}, \cite{Hinton_2018} use a dynamic routing algorithm to calculate a set of routing coefficients that link lower and higher-level capsules between adjacent layers in the network. Each routing coefficient represents the probability that an individual lower-level capsule should be assigned to a higher-level capsule. These routing coefficients are not learned during training, as is the case for the rest of the network parameters. For each input presented to the network, the routing coefficients are calculated at run-time (during both training and inference) from a set of initial values.

The Softmax function, given by Eq. \ref{Eq: Softmax Normalization}, has been widely used for object recognition tasks due to its ability to reduce the impact of outlier values in the dataset while still allowing those values to have an effect on the network's learning ability during training. In CapsNets, Softmax is used to convert the log priors between capsules \textit{i} in layer \textit{l} and capsules \textit{j} in layer \textit{l + 1} into a set of assignment probabilities between the capsules. While dampening the effects of outliers can be beneficial when training typical network parameters (following the Maximum Likelihood Estimation principle), outliers in the routing coefficients can provide optimal separation between features in adjacent capsule layers. Since these coefficients are not learned in the conventional sense (i.e., gradients do not flow through the routing coefficients during backpropagation), other normalization functions and methods can be used for the task of dynamic routing. In addition, the function need not be differentiable (e.g., a look-up table can be used to assign lower-level capsules to higher-level capsules).

Here, we show that the use of the scale-invariant Max-Min function (Eq. \ref{Eq: Max-Min Normalization}) improves the performance of CapsNets. We focus on the CapsNet formalism of Sabour et al. \cite{Sabour_2017}. The lower bound for the normalization is set to $0.0$ and gives higher-level capsules the ability to completely disregard non-essential features presented by one of the lower-level capsules. This serves as a kind of dynamic dropout for the routing coefficients and forces the network to generalize better. The upper bound, in principle, can be set to any value. We tested upper bounds in the range of $0.01 - 1.0$ and found that the network performs well for all values within the range (we use $1.0$ as the upper bound in the rest of the paper). Bounding the routing coefficients between $0.0$ and $1.0$ in this manner allows each lower-level capsule to have an independent assignment probability to each of the higher-level capsules. That is, the sum of the probabilities for a single lower-level capsule across each of the higher-level capsules is no longer constrained to be $1.0$. This can be beneficial for CapsNets since, often times, a single feature might have high probabilities of being assigned to multiple higher-level objects.

The use of Max-Min over Softmax leads to an improvement in the test accuracy across five datasets and allows the use of more routing iterations between capsule layers without overfitting to the training data. In addition, we train a single CapsNet (with minimal data augmentation) on the MNIST dataset and achieve a test error of $0.2\%$. With a $3$-model majority voting system, we achieve a test error of $0.17\%$ on MNIST, surpassing the accuracy of the model ensemble used by \cite{Wan_2013} by $19\%$.

Section \ref{Section: Capsule Network Architecture} provides a summary of the three-layer CapsNet from \cite{Sabour_2017} and the differences in the routing procedure between the Softmax and Max-Min normalizations. In Section \ref{Section: Evolution of Logits and Routing Coefficients}, we compare the evolution of the logits and routing coefficients for a CapsNet trained using Softmax and Max-Min. Section \ref{Section: DigitCaps Outputs} shows the tuning curves (i.e., outputs of the routing layer) for the network. Section \ref{Section: Results} shows the main results of the sessions trained using Softmax vs. Max-Min. Performance on the MNIST dataset is detailed in Section \ref{Section: Performance on MNIST} along with results on the MNIST dataset using other normalization functions.

\section{Capsule Network Architecture}
\label{Section: Capsule Network Architecture}
\begin{figure}[h]
\centering
{\includegraphics[width = 5.5 in]{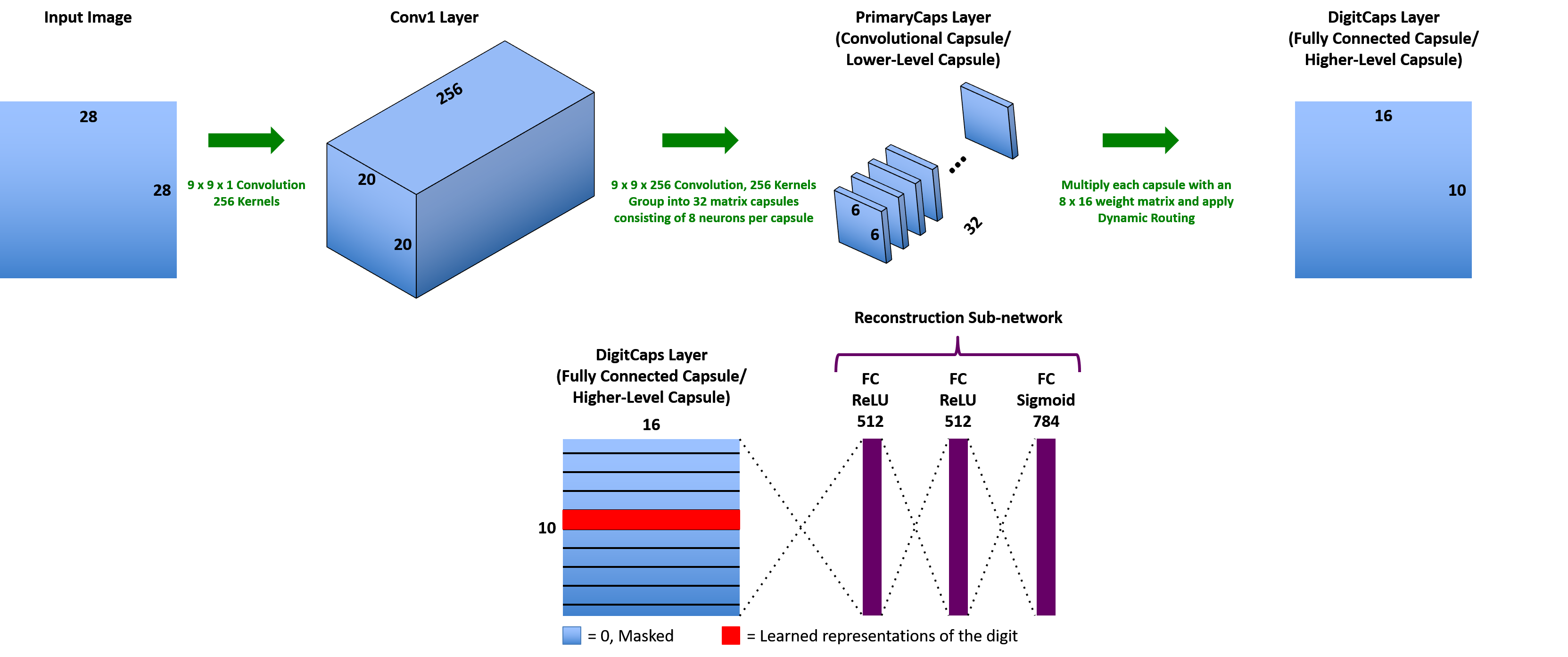}}
\caption{(Top) Three-layer CapsNet architecture, following Sabour et al. \cite{Sabour_2017}. PrimaryCaps layer consists of $6\times6\times32$ $=$ $1152$ $8$-D vector capsules. The dynamic routing algorithm produces the $10\times16$ DigitCaps layer, which is used to calculate the margin loss. (Bottom) Reconstruction sub-network with three fully-connected layers. The DigitCaps layer is passed to the reconstruction sub-network where the non-ground-truth class rows are masked with zeros. The sparse $10\times16$ matrix is then passed through the network, which learns to reproduce the input image. Margin and reconstruction loss functions follow those from \cite{Sabour_2017}.}
\label{Fig: CapsNet Architecture}
\end{figure}

The CapsNet architecture we used follows the network described in \cite{Sabour_2017} and is shown in Fig. \ref{Fig: CapsNet Architecture}. A $28\times28$ input image is fed into a convolutional layer (Conv1) that operates on the input with $256$, $9\times9$ kernels using a stride of $1$ and the ReLU activation. The output of this operation is a $20\times20\times256$ feature map tensor that is then fed into a second convolutional layer (PrimaryCaps) that uses $256$, $9\times9$ kernels, a stride of $2$ and the ReLU activation. This results in a $6\times6\times256$ feature map tensor, which represents the lower-level capsules for the network. Each set of $8$ scalar neurons in the $6\times6\times256$ tensor is grouped channel-wise and forms a single lower-level capsule $i$, for a total of $6\times6~\times$ ($256\div8$) $=1152$ lower-level capsules.

The outputs from PrimaryCaps are fed through a dynamic routing algorithm, resulting in the DigitCaps output matrix. The squashing function used to calculate \boldmath$v_j$ is as given in \cite{Sabour_2017}. Each row in the DigitCaps matrix represents the $16$-D instantiation parameters of a single class and the length of a $16$-D vector represents the probability of the existence of a particular class. During training, the non-ground-truth rows are masked with zeros and the matrix is passed to a reconstruction sub-network that consists of two fully-connected layers of dimensions $512$ and $1024$ with ReLU activations and a final fully-connected layer of dimension $784$ with a sigmoid activation. During inference, the row in the DigitCaps matrix with the largest length (i.e., highest probability) is taken as the predicted object class.

The inputs to the routing algorithm consist of the prediction vectors, \boldmath$\hat{u}_{j|i}$. These prediction vectors are calculated using learned transformation weight matrices and the capsule outputs from the PrimaryCaps layer. The prediction vectors remain fixed inside the algorithm as the routing procedure boot-strap calculates the DigitCaps capsules, \boldmath$v_j$, using the prediction vectors. Although there are no gradient flows in the routing layer, both the inputs and outputs of the routing layer are subjected to the usual gradient flows during training. In particular, the DigitCaps capsules are passed onto a sub-network that learns to reconstruct the original input image. As a result, the prediction vectors and parent-level capsules tend to evolve such that the scaled summation of the prediction vectors are similar to the parent-level capsules. In other words, during the forward pass, the network calculates a set of parent-level capsules that are used to recreate the original image. Any errors in the reconstruction network will backpropagate themselves to the prediction vectors and the preceding layers. During the next forward pass, the prediction vectors will evolve in such a way (via the transformation matrices) that aligns with the previously calculated parent-level capsules. 

The routing procedure from \cite{Sabour_2017} is given below for reference. The routing procedure using Max-Min normalization remains largely the same except the Softmax function is replaced with the Max-Min function as given by Eq. \ref{Eq: Max-Min Normalization}, where $p$/$q$ are the lower/upper bounds of the normalization. For the first iteration, the \textit{routing coefficients} are initialized to $1.0$ outside of the routing for-loop.

Training is conducted similar to the approach taken in \cite{Sabour_2017}. Our implementation uses TensorFlow \cite{TensorFlow} and the Adam optimizer \cite{Adam_Optimizer} with TensorFlow's default parameters and an exponentially decaying learning rate. Unless otherwise noted, the same network hyperparameters in \cite{Sabour_2017} were used for all training sessions. Original code is adapted from \cite{Sabour_Code}.

\begin{table}[h!]
    \begin{tabular}{l}
    \hline
    \textbf{Softmax Routing Procedure} \\
    \hline
    1: Input to Routing Procedure: ($\bm{\hat{u}_{j|i}}$, $r$, $l$) \\
    2: \quad for all capsules $i$ in layer $l$ and capsule $j$ in layer ($l$ + 1): $b_{ij}$ $\leftarrow$ 0 \\
    3: \quad \textbf{for} $r$ iterations: \\
    4: \quad \quad for all capsule $i$ in layer $l$: $\bm{c_i}$ $\leftarrow$ Softmax($\bm{b_i}$) \\
    5: \quad \quad for all capsule $j$ in layer ($l$ + 1): $\bm{s_j}$ $\leftarrow$ $\sum_{i} c_{ij} \bm{\hat{u}_{j|i}}$ \\
    6: \quad \quad for all capsule $j$ in layer ($l$ + 1): $\bm{v_j}$ $\leftarrow$ Squash(\bm{$s_j$}) \\
    7: \quad \quad for all capsule $i$ in layer $l$ and capsule $j$ in layer ($l$ + 1): $b_{ij} \leftarrow b_{ij} + \bm{\hat{u}_{j|i}} \cdot$ $\bm{v_j}$ \\
       \quad \quad \textbf{return} $\bm{v_j}$ \\
    \hline
    \end{tabular}
    \label{Procedure: Softmax Routing}
\end{table}

\begin{equation}
	\label{Eq: Softmax Normalization}
	c_{ij} = \frac{exp(b_{ij})}{\sum_k exp(b_{ik})}
\end{equation}

\begin{table}[h!]
    \begin{tabular}{l}
    \hline
    \textbf{Max-Min Routing Procedure} \\
    \hline
    1: Input to Routing Procedure: ({$\bm{\hat{u}_{j|i}}$}, $r$, $l$) \\
    2: \quad for all capsules $i$ in layer $l$ and capsule $j$ in layer ($l$ + 1): $c_{ij}$ $\leftarrow$ 1.0 \\
    3: \quad \textbf{for} $r$ iterations: \\
    4: \quad \quad for all capsule $j$ in layer ($l$ + 1): $\bm{s_j}$ $\leftarrow$ $\sum_{i} c_{ij} \bm{\hat{u}_{j|i}}$ \\
    5: \quad \quad for all capsule $j$ in layer ($l$ + 1): $\bm{v_j}$ $\leftarrow$ Squash($\bm{s_j}$) \\
    6: \quad \quad for all capsule $i$ in layer $l$ and capsule $j$ in layer ($l$ + 1): $b_{ij} \leftarrow b_{ij} + \bm{\hat{u}_{j|i}} \cdot$ $\bm{v_j}$ \\
    7: \quad \quad for all capsule $i$ in layer $l$: $\bm{c_i}$ $\leftarrow$ Max-Min ($b_{ij}$) $\Longrightarrow$ $\mathrm{Given~in~Eq.}$ ~\ref{Eq: Max-Min Normalization} \\
       \quad \quad \textbf{return} $\bm{v_j}$ \\
    \hline
    \end{tabular}
    \label{Procedure: Max-Min Routing}
\end{table}

\begin{equation}
	\label{Eq: Max-Min Normalization}
	c_{ij} = p + \frac{b_{ij} - min(b_{ij})}{max(b_{ij}) - min(b_{ij})} * (q-p)
\end{equation}

\section{Evolution of Logits and Routing Coefficients}
\label{Section: Evolution of Logits and Routing Coefficients}
In the CapsNet architecture presented in Fig. \ref{Fig: CapsNet Architecture}, the capsules in the PrimaryCaps layer can represent features useful for recognizing objects (e.g., hand-written digits). The convolutional layer before PrimaryCaps allow efficient learning of these features. The dynamic binding of parts (i.e., features) to wholes (i.e., objects) are then carried out through the routing coefficients.

\begin{figure}[h]
\centering
{\includegraphics[width = 5.5 in]{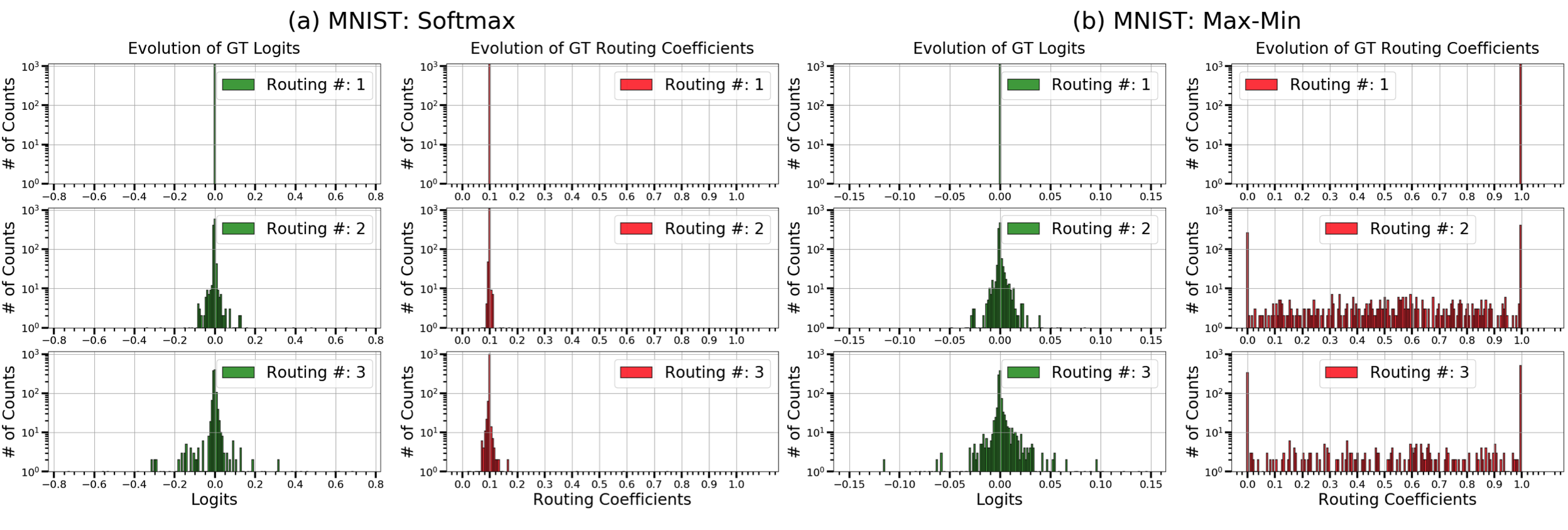}}
\caption{Evolution of logits and routing coefficients for the same training image from the MNIST dataset for networks trained using (a) Softmax and (b) Max-Min. Y-axes are displayed in log-scale. Note: For clarity, only the routing coefficients associated with the ground-truth (GT) column are shown here. The histograms of all routing coefficients exhibit the same behavior.}
\label{Fig: Evolution of Logits and Routing Coefficients MNIST}
\end{figure}

\begin{figure}[h]
\centering
{\includegraphics[width = 5.5 in]{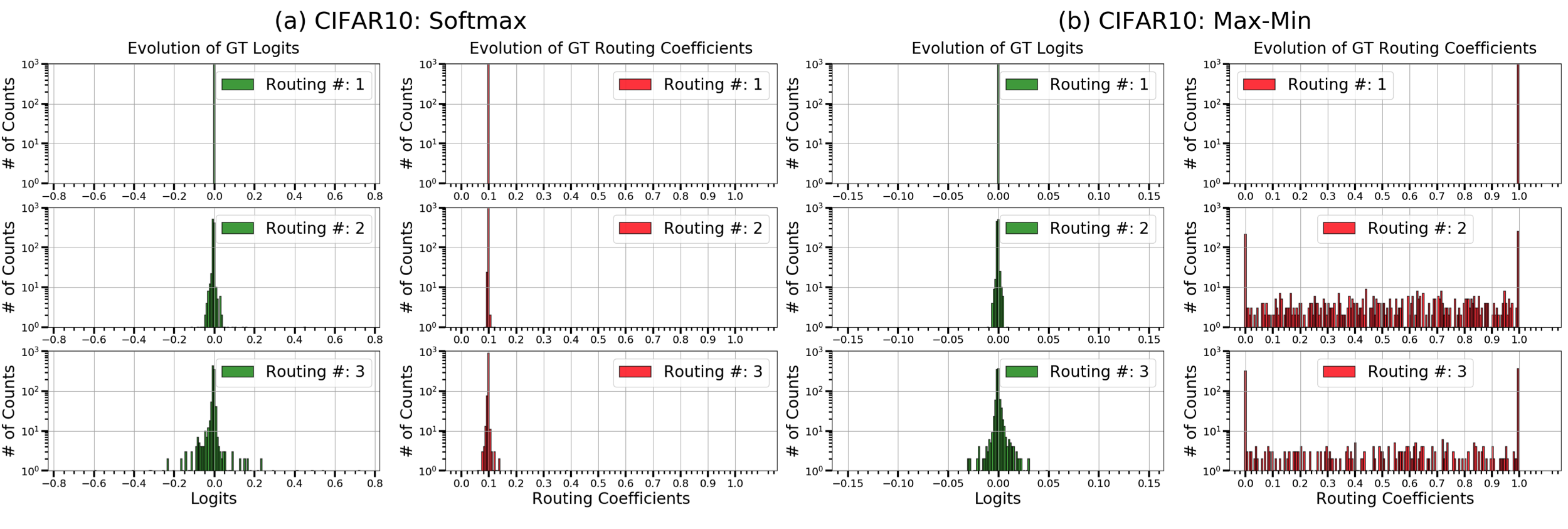}}
\caption{Evolution of logits and routing coefficients for the same training image from the CIFAR dataset for networks trained using (a) Softmax and (b) Max-Min. Y-axes are displayed in log scale. Note: For clarity, only the routing coefficients associated with the GT column are shown here. The histograms of all routing coefficients exhibit the same behavior.}
\label{Fig: Evolution of Logits and Routing Coefficients CIFAR10}
\end{figure}

In such CapsNets, the ability to create optimal separation between competing features in adjacent capsule layers is a crucial feature for efficient object recognition. The evolution of the logits and routing coefficients in the routing layer of a CapsNet offer insights into how two adjacent capsule layers assign object features to their wholes. When using Softmax, a zero initialization for the logits sets the routing coefficients to be $0.1$ (assuming $10$ parent-level output capsules) for the first iteration. With Max-Min normalization, the routing coefficients are initialized to $1.0$ for the first iteration; thus, for the first iteration, the parent-level capsules, $\bm{v_{j}}$, are simply the (non-scaled) squashed summation of the prediction vectors.
The top rows of Figs. \ref{Fig: Evolution of Logits and Routing Coefficients MNIST} and \ref{Fig: Evolution of Logits and Routing Coefficients CIFAR10} show the initial values of the logits and routing coefficients for MNIST and CIFAR10, each for the same training image in their respective datasets. The middle and bottom rows show the evolution of the logits and coefficients throughout the routing procedure. For both the Softmax and Max-Min cases, the logits and coefficients are extracted from the network after their respective sessions have finished training under the same conditions.

As the routing progresses, the logits form a tight cluster around zero, with the majority of the values remaining at $0.0$ (y-axes are log-scale for Figs. \ref{Fig: Evolution of Logits and Routing Coefficients MNIST} and \ref{Fig: Evolution of Logits and Routing Coefficients CIFAR10}). Due to the tight clustering and the non-linear behavior of Softmax, the routing coefficients from a Softmax trained network evolves in a manner similar to their corresponding logits (c.f. Figs. \ref{Fig: Evolution of Logits and Routing Coefficients MNIST} (a) and \ref{Fig: Evolution of Logits and Routing  Coefficients CIFAR10} (a)); i.e., the majority of the routing coefficients remain at their initial value of $0.1$, with only a few that evolve to significantly different values. As a result, the routing coefficients just barely separate each lower-level capsule among the higher-level capsules. With Max-Min normalization, the majority of the logits also have a value of $0.0$. However, due to the scale-invariant nature of the Max-Min normalization, the tight grouping of logits can be better separated to form the routing coefficients (c.f. Figs. \ref{Fig: Evolution of Logits and Routing Coefficients MNIST} (b) and \ref{Fig: Evolution of Logits and Routing Coefficients CIFAR10} (b)).

\begin{figure}[h]
\centering
{\includegraphics[width = 5.5 in]{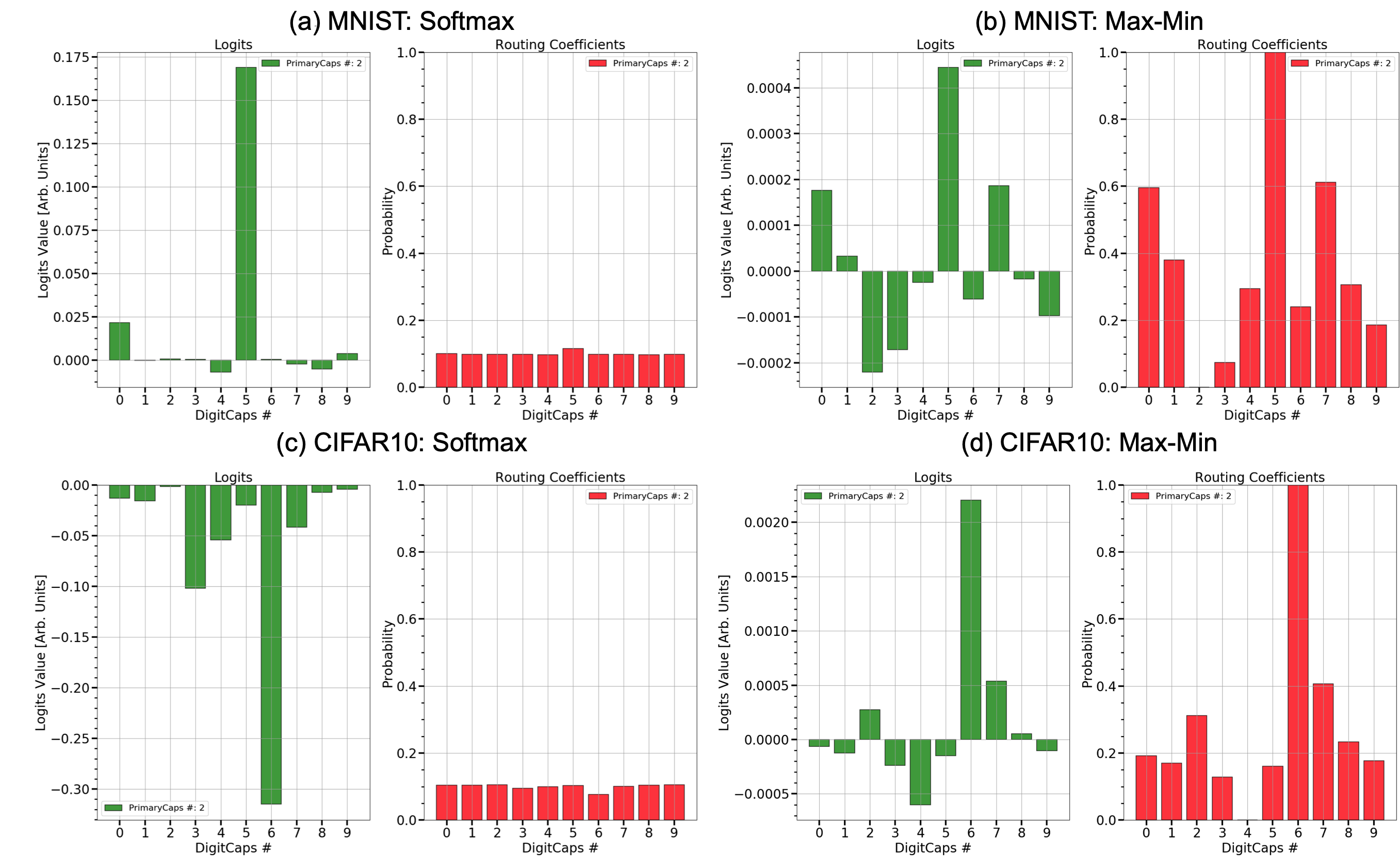}}
\caption{Examples of logits and corresponding routing coefficients for an individual lower-level capsule from PrimaryCaps for the MNIST (top row) and CIFAR10 (bottom row) datasets for networks trained using Softmax ((a) and (c)) and Max-Min ((b) and (d)) normalizations at the last routing iteration.}
\label{Fig: Single Row Logits and Routing Coefficients MNIST CIFAR10}
\end{figure}

Max-Min also allows a lower-level capsule to have high assignment probabilities with multiple higher-level capsules. With Softmax, competition between a lower-level capsule and each of the higher-level capsules reduces the likelihood of multiple high probabilities between features in adjacent capsule layers. Figure \ref{Fig: Single Row Logits and Routing Coefficients MNIST CIFAR10} shows examples of the routing coefficients for three lower-level capsules in PrimaryCaps across the ten higher-level capsules in DigitCaps for the MNIST and CIFAR10 datasets at the last routing iteration. Since the majority of the logits are tightly clustered around their initial values, Softmax computes nearly identical assignment probabilities between a lower-level capsule and each of the higher-level capsules. Max-Min normalization computes high assignment probabilities for multiple higher-level capsules. In addition, the differences between the probabilities among the higher-level capsules (for each lower-level capsule) is larger when Max-Min is used. This leads to a more optimal separation between capsules in adjacent layers.

\section{DigitCaps Outputs}
\label{Section: DigitCaps Outputs}
It is also instructive to examine the outputs of the DigitCaps layer (i.e., the parent-level capsules, \boldmath$v_j$). During inference, the output capsule with the largest vector length (i.e., highest probability) is used to classify the input image. For each input to the network, an ideal CapsNet would have a single output capsule with probability near $1.0$ corresponding to the GT class and all other classes with probabilities near $0.0$. The outputs can be examined on an input-by-input basis (for each image, there is a corresponding $10\times16$ DigitCaps matrix from which the classification is made) or on a class-by-class basis (for each class, there is a corresponding $10\times16$ matrix that is the average of the individual matrices for that class).

\begin{figure}[h]
\centering
{\includegraphics[width = 5.5 in]{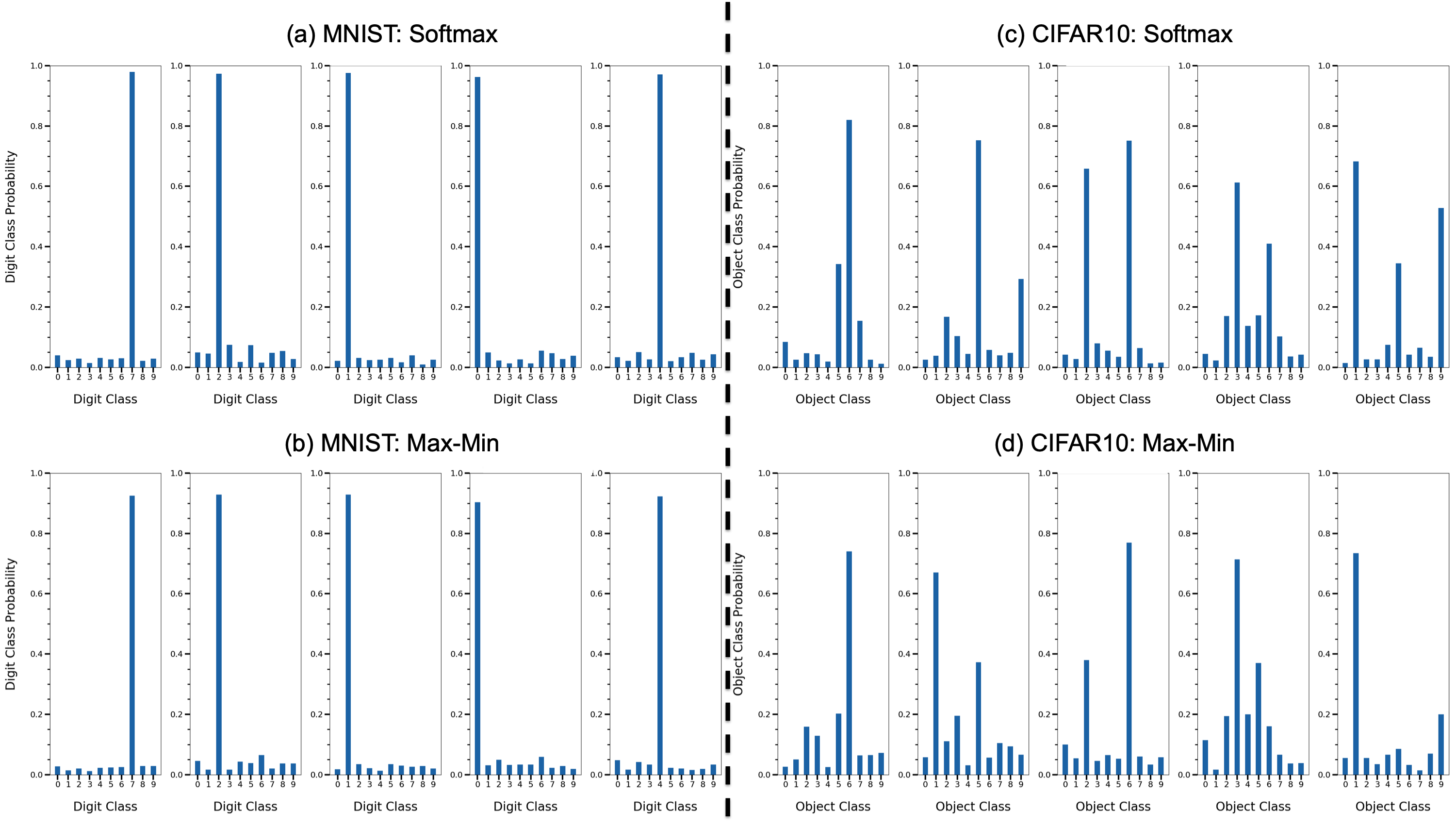}}
\caption{Output class probabilities for the same set of test images from the MNIST and CIFAR10 datasets calculated using Softmax ((a) and (c)) and Max-Min ((b) and (d)) normalizations.}
\label{Fig: v_Js MNIST and CIFAR10}
\end{figure}

Figures \ref{Fig: v_Js MNIST and CIFAR10} (a) and (b) show the output capsule probabilities for the same set of test images from the MNIST dataset and Figs. \ref{Fig: v_Js MNIST and CIFAR10} (c) and (d) show the output capsule probabilities for the same set of test images from the CIFAR10 dataset. For MNIST, the network is properly trained and both normalizations provide digit class probabilities that are highly-peaked for their corresponding GT classes and vice versa for the other classes. For CIFAR10, the network does better at separating the classes when Max-Min normalization is used (i.e., higher GT probability and lower non-ground-truth probabilities). However, both normalizations produce output capsules that have multiple high peaks, signifying that the networks are not able to adequately differentiate between the object classes. This issue with CapsNets was addressed in \cite{Sabour_2017} by including a ``none-of-the-above'' category for the routing Softmax---our network does not have this category.

\begin{figure}[h]
\centering
{\includegraphics[width = 5.5 in]{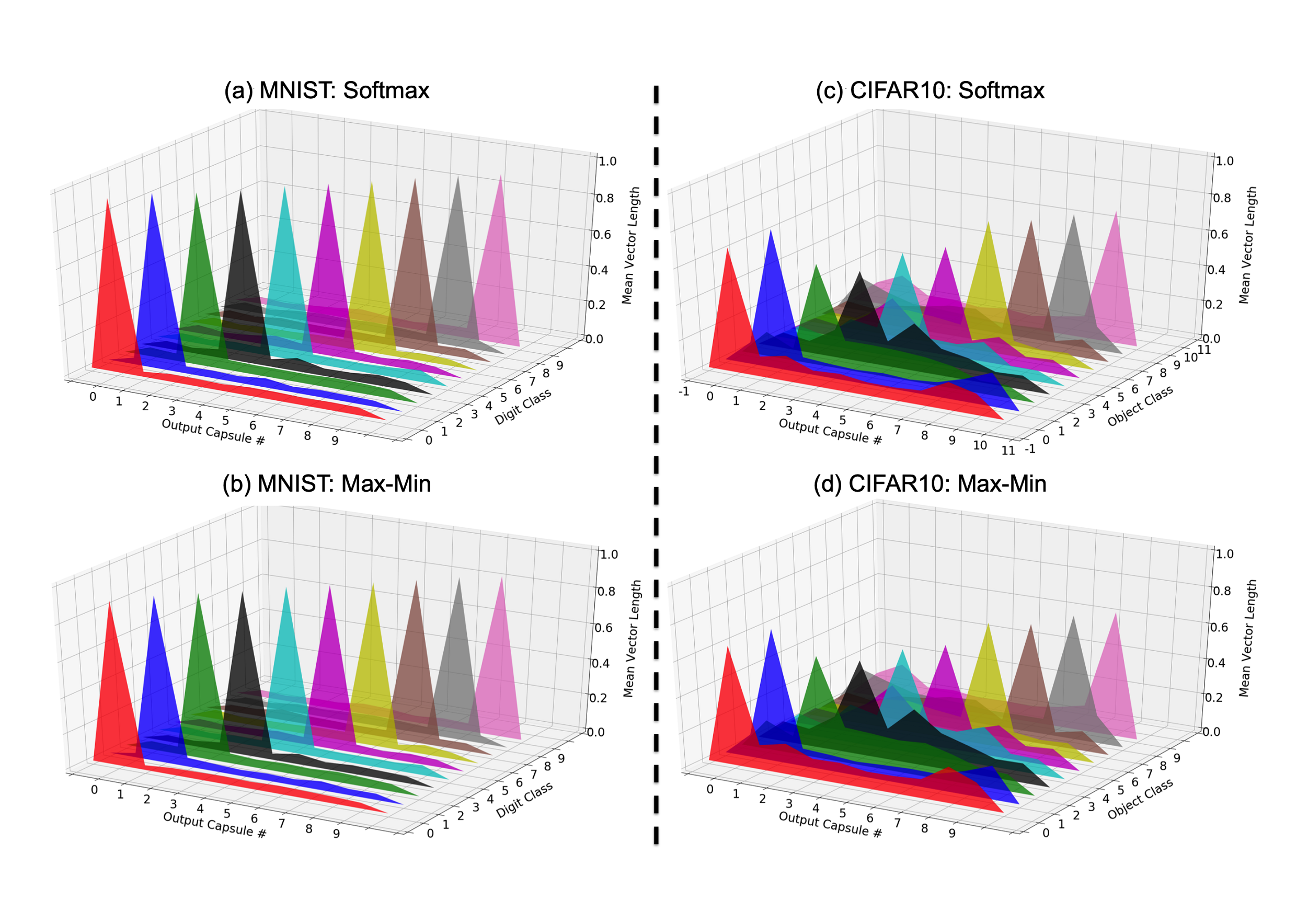}}
\caption{Class-averaged output probabilities for the MNIST and CIFAR10 test datasets calculated using Softmax ((a) and (c)) and Max-Min ((b) and (d)) normalizations. The shape similarities in the tuning curves between the Softmax and Max-Min trained networks suggests that both models are trained in a similar fashion.}
\label{Fig: Tuning Curves MNIST and CIFAR10}
\end{figure}

If we view the capsules in the DigitCaps layer to be “grandmother cells” \cite{Gross_2002}, then how well-tuned they are to the objects they recognize provides a picture about the robustness of the system. Figures \ref{Fig: Tuning Curves MNIST and CIFAR10} (a) and (b) show the class-averaged output capsule probabilities for the test images for the MNIST dataset. These can be viewed as the tuning curves of the recognition units and demonstrate the ability of the network to adequately distinguish between each of the digit classes. The similarity between the tuning curves of the respective digits when trained with Softmax (Fig. \ref{Fig: Tuning Curves MNIST and CIFAR10} (a)) and Max-Min (Fig. \ref{Fig: Tuning Curves MNIST and CIFAR10} (b)) show that Max-Min normalization does not degrade the network’s ability to discriminate the ten digits. In contrast, the tuning curves for CIFAR10 (Figs. \ref{Fig: Tuning Curves MNIST and CIFAR10} (c) and (d)) show that, for certain object classes, the discriminability is not as good as those for MNIST. This is also reflected in the accuracies given in Table \ref{Table: Test Accuracies on Datasets}.

\section{Results}
\label{Section: Results}
We compared the network's performance with Max-Min and Softmax normalizations on five datasets: MNIST \cite{MNIST}, Background MNIST (bMNIST) and Rotated MNIST (rMNIST) \cite{R_and_BG_MNIST}, Fashion MNIST (fMNIST) \cite{F_MNIST}, and CIFAR10 \cite{CIFAR10}. In addition, we also evaluated the performance of the networks as a function of the number of routing iterations. All sessions were trained using the same three-layer model shown in Fig. \ref{Fig: CapsNet Architecture} and hyperparameters as in \cite{Sabour_2017}. For the results in this section, no data augmentations were used for the datasets with the exception of CIFAR10, where random $24\times24$ croppings were conducted for the training images and a centered $24\times24$ cropping conducted for the test images. For variations of the MNIST dataset, the PrimaryCaps layer has $32$ capsules. For CIFAR10, the PrimaryCaps layer has $64$ capsules. Three routing iterations were used for all sessions. Unlike \cite{Sabour_2017}, we did not introduce a ``none-of-the-above'' category for the network classifier.

\begin{figure}[h]
\centering
{\includegraphics[width = 5.5 in]{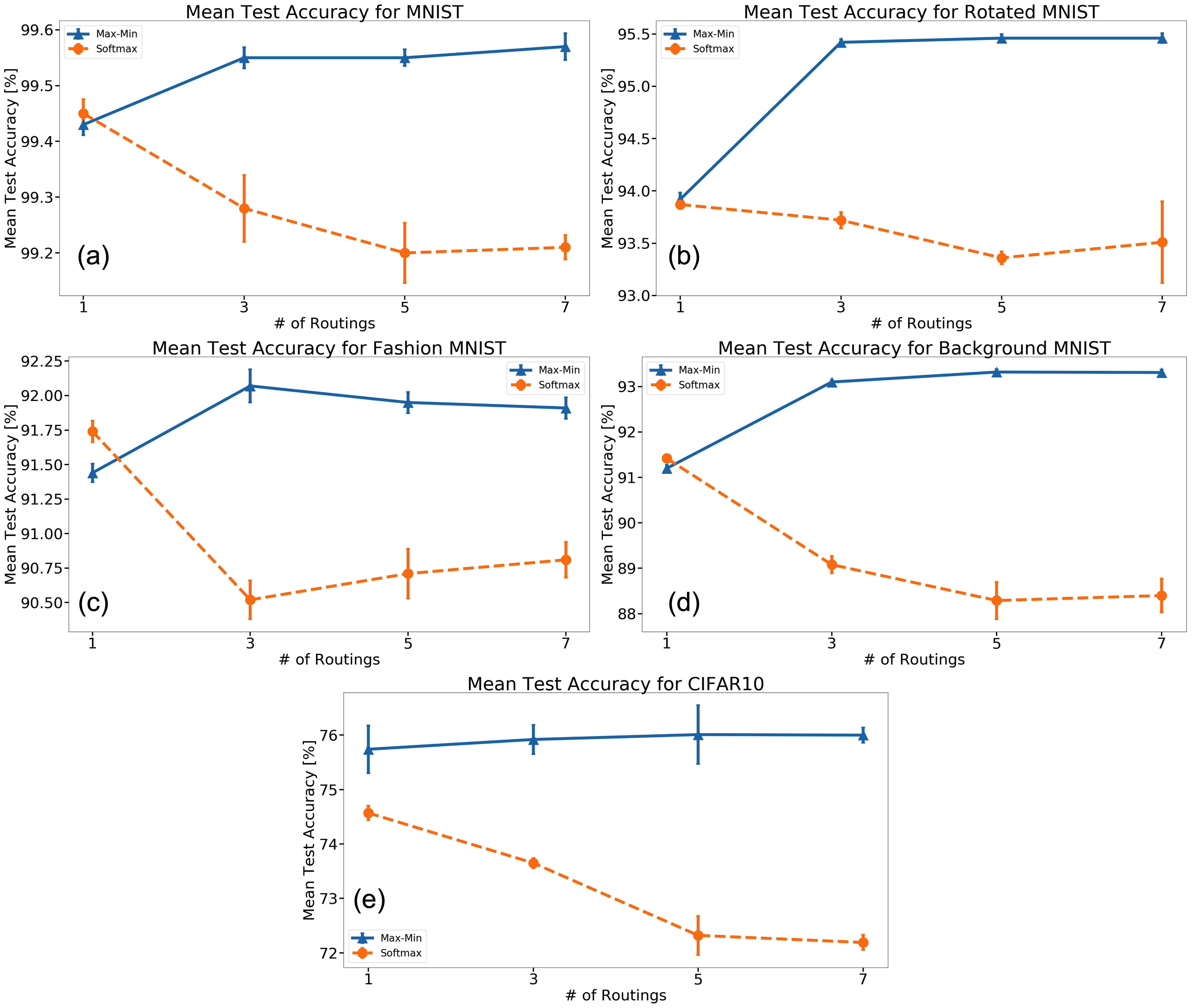}}
\caption{Test accuracy as a function of the number of routing iterations for CapsNets trained using Max-Min (blue solid lines with triangle markers) and Softmax (orange dashed lines with circle markers) normalization for (a) MNIST, (b) Rotated MNIST, (c) Fashion MNIST, (d) Background MNIST, and (e) CIFAR10. Max-Min normalization allows the network capacity to increase without decreasing the network performance.}
\label{Fig: Test Accuracy vs. NumRouting}
\end{figure}

Table \ref{Table: Test Accuracies on Datasets} lists the mean of the maximum test accuracies and their standard deviations for the five datasets, and shows that Max-Min normalization provides a consistent improvement in test accuracy compared with Softmax.\footnote{Experiments using Softmax normalization and routing coefficients initialized to $1.0$ produced test accuracies of $99.30\pm0.03\%$ for the MNIST dataset.} In particular, the improvement is most significant for datasets that have a non-zero background (i.e., bMNIST and CIFAR10). Max-Min also allows more routing iterations to be conducted without decreasing the test accuracy. As shown in Fig. \ref{Fig: Test Accuracy vs. NumRouting}, an improvement in test accuracy is obtained when the number of routing iterations is increased. With Softmax, the test accuracy decreases for all five datasets.

\begin{table}[h]
    \centering
    \caption{Mean of the maximum test accuracies and their standard deviations on five datasets for Max-Min and Softmax normalizations. Five training sessions were conducted for each dataset for both Max-Min and Softmax. For CIFAR10, a ``none-of-the-above'' category was \textit{not} used during training.}
    \begin{tabular}{cccccc}
    \hline
    \textbf{Normalization} & \textbf{MNIST [$\%$]} & \textbf{rMNIST [$\%$]} & \textbf{fMNIST [$\%$]} & \textbf{bMNIST [$\%$]} & \textbf{CIFAR10 [$\%$]}  \\
    \hline
    Softmax & 99.28 $\pm$ ~0.06 & 93.72 $\pm$ ~0.08 & 90.52 $\pm$ ~0.14 & 89.08 $\pm$ ~0.19 & 73.65 $\pm$ ~0.09  \\
    Max-Min & \textbf{99.55} $\pm$ ~0.02 & \textbf{95.42} $\pm$ ~0.03 & \textbf{92.07} $\pm$ ~0.12 & \textbf{93.09} $\pm$ ~0.04 & \textbf{75.92} $\pm$ ~0.27 \\
    \hline
    \end{tabular}
    \label{Table: Test Accuracies on Datasets}
\end{table}

\begin{figure}[h]
\centering
{\includegraphics[width = 5.5 in]{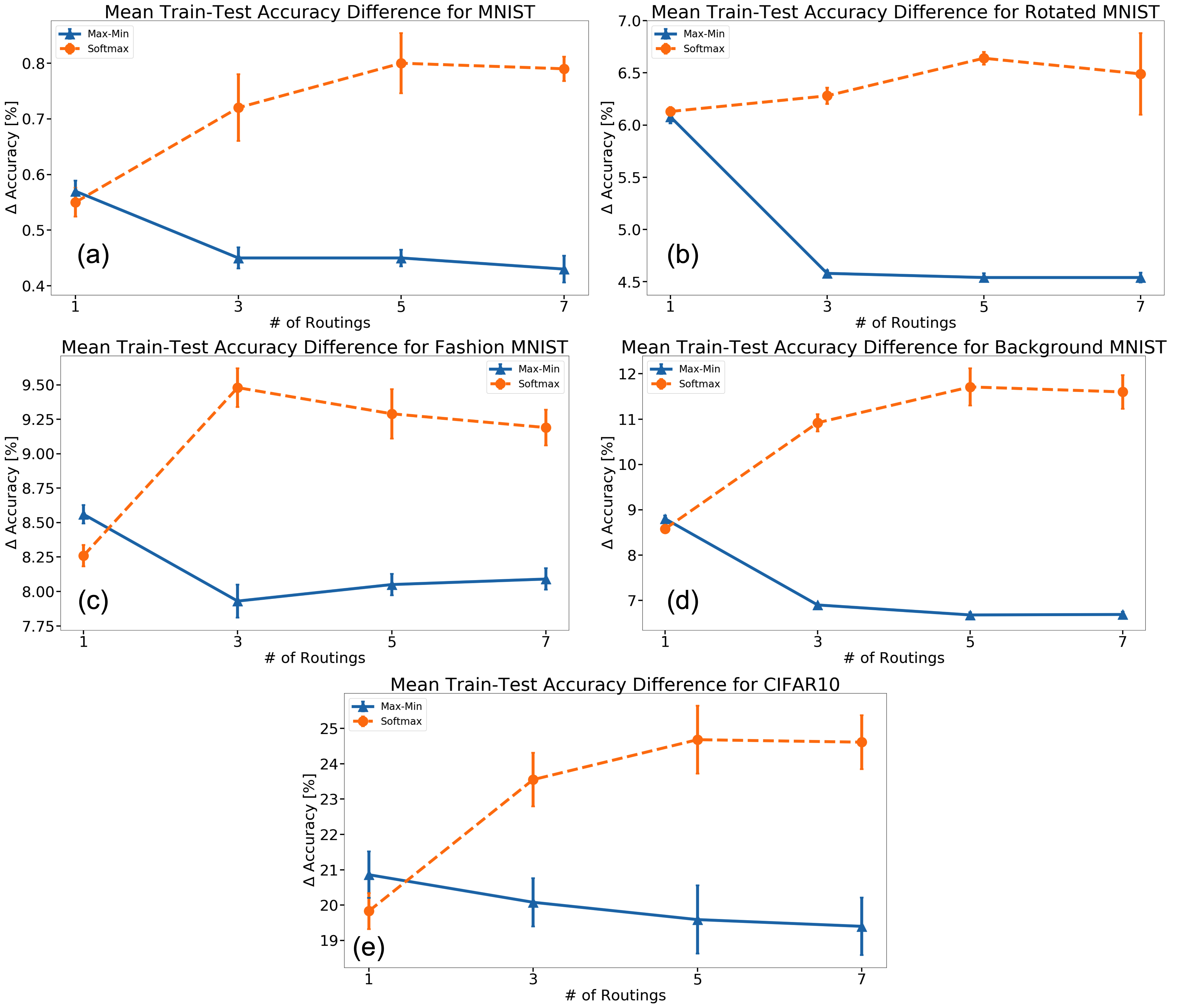}}
\caption{Difference between training and test accuracies as a function of the number of routing iterations for CapsNets trained using Max-Min (blue solid lines with triangle markers) and Softmax (orange dashed lines with circle markers) normalization for (a) MNIST, (b) Rotated MNIST, (c) Fashion MNIST, (d) Background MNIST, and (e) CIFAR10. Max-Min normalization prevents the network from overfitting on the training data, especially as the number of routing iterations is increased.}
\label{Fig: Delta Accuracy vs. NumRouting}
\end{figure}

Max-Min also prevents the network from overfitting to the training data, especially as the number of routing iterations is increased. As shown in Fig. \ref{Fig: Delta Accuracy vs. NumRouting}, the differences between the mean of the maximum training and test accuracies are lower for CapsNets trained using Max-Min compared with Softmax. Thus, Max-Min normalization not only prevents the model from overfitting, but also allows the performance of the network to scale positively with the number of routing iterations.

\section{Performance on MNIST}
\label{Section: Performance on MNIST}
Sabour et al. demonstrates a low test error of $0.25\%$ \cite{Sabour_2017} using a single three-layer CapsNet with Softmax and performing image translation by up to $2$ pixels in each direction with zero padding. Section \ref{Section: Results} shows that Max-Min gives an average of $0.27\%$ improvement in test accuracy compared with Softmax for MNIST. Thus, it stands to reason that a single CapsNet trained using Max-Min and minimal augmentations can outperform the current state-of-the-art results on MNIST \cite{Wan_2013}. We train the same three-layer CapsNet in Fig. \ref{Fig: CapsNet Architecture} on the full $60,000$ $28\times28$ MNIST training images using random image translation by up to $2$ pixels with zero padding and random image rotation (around image center) by up to $20$ degrees. In addition, we relax the margin loss constraints such that $m^+=0.8$ and $m^-=0.2$ and use only $3$ routing iterations. The batch size used for training is $400$ images and the networks are trained for $500$ epochs each. All other parameters follow those from \cite{Sabour_2017}.

Table \ref{Table: Performance on MNIST} gives a comparison on the test errors for the $10,000$ images in the MNIST test dataset for networks trained with Max-Min and Softmax using the parameters and image augmentations listed above. Each experiment was conducted a total of $10$ times. A single CapsNet using Max-Min achieves a test error of $0.20\%$, while a 3-model majority vote achieves a test error of $0.17\%$. The $17$ misclassified images from the model ensemble are shown in Fig. \ref{Fig: Ensemble Misclassifications}. Further discussions on the MNIST results are presented in Appendix \ref{App. Misclassified Images From the MNIST Dataset} along with misclassifications from each of the three models used in the ensemble.

\begin{table}[h]
    \centering
    \caption{Test errors on MNIST dataset for Max-Min and Softmax normalizations. Ten training sessions were conducted for both the Max-Min and Softmax normalizations.}
    \begin{tabular}{ccccc}
    \hline
    \textbf{Normalization} & \textbf{Maximum} & \textbf{Minimum}    & \textbf{Mean} & \textbf{Stdev.}  \\
    \hline
    Softmax                & 0.35\%       & 0.29\%          & 0.32\%        & 0.021\%  \\
    Max-Min                & 0.27\%       & \textbf{0.20\%} & 0.24\%        & 0.025\%  \\
    \hline
    \end{tabular}
    \label{Table: Performance on MNIST}
\end{table}

\begin{figure}[h]
\centering
{\includegraphics[width = 5.5 in]{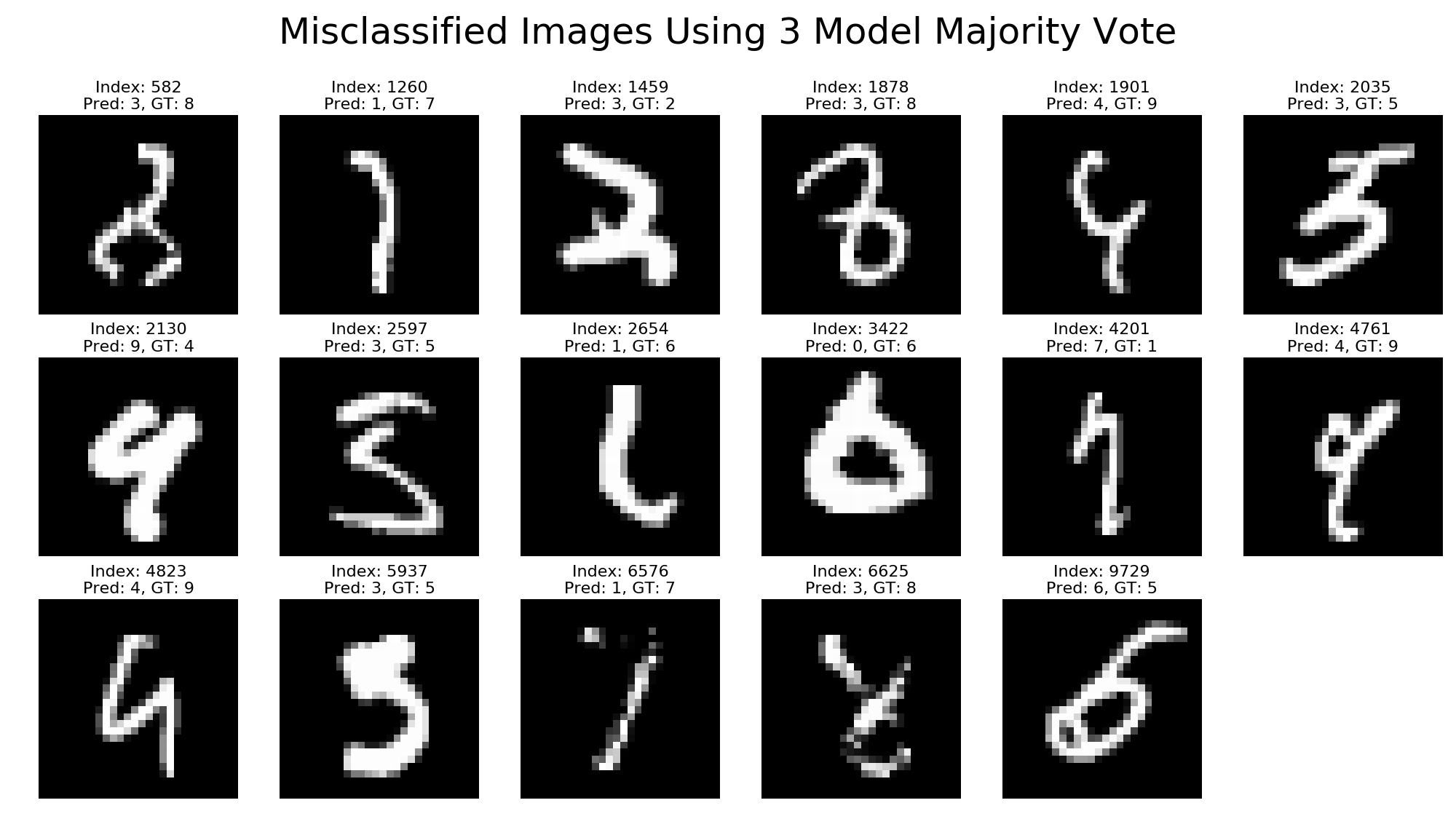}}
\caption{Misclassified MNIST images using 3-model majority vote from CapsNets trained using Max-Min normalization.}
\label{Fig: Ensemble Misclassifications}
\end{figure}

\subsection{Comparisons with Human Polling Results}
\label{Section: Comparisons with Human Polling Results}
\begin{figure}[h]
\centering
{\includegraphics[width = 3.0 in]{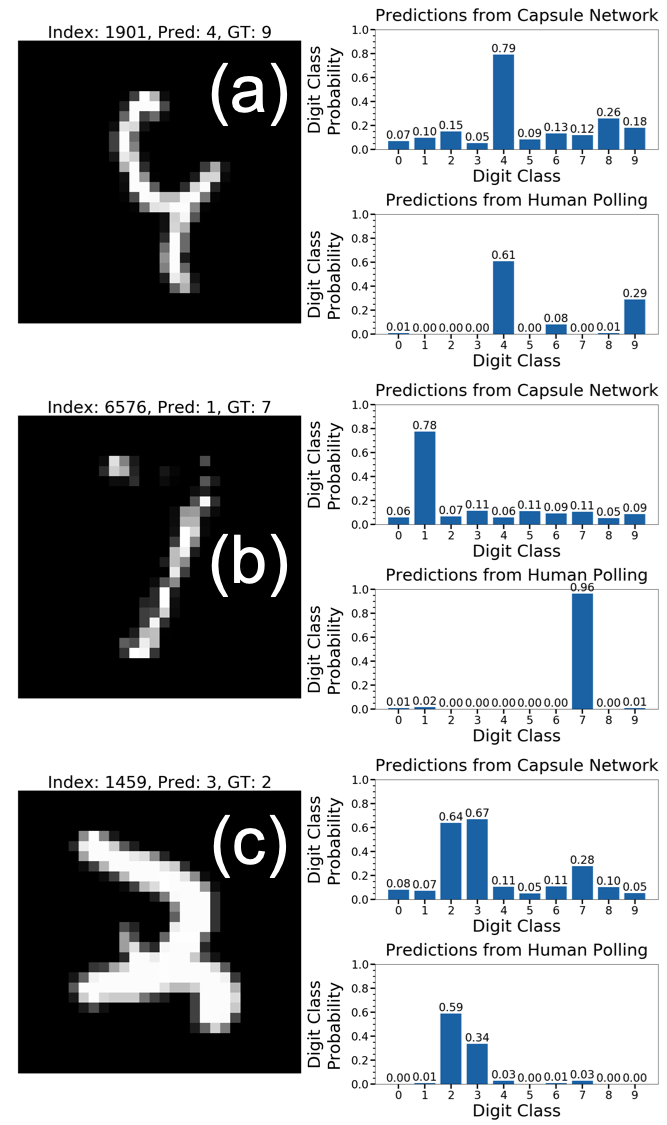}}
\caption{(a) Agreement between human and network predictions for a misclassified image. Here, the majority of the human polling results agree with the network predictions and both disagree with the GT label. (b) Disagreement between human and network predictions for a misclassified image. Here, the majority of the human polling results agree with the GT label. (c) Example of human and network predictions for a poor quality image. Although the majority of the human predictions agree with the GT label, a significant portion misclassified the image as the digit $3$. Note: The predictions from human polling are averaged over $120$ individuals who took the same poll whereas the network predictions are averaged over three separately trained models (and hence, the probabilities do not sum to $1.0$).}
\label{Fig: Human-Network Polling Plots}
\end{figure}

Out of $10,000$ predictions the network makes on the MNIST test set, $17$ differ from the GT labels. A poll of $120$ individuals on these $17$ images showed that in some cases ($6$ out of $17$) the results agreed with the networks’ predictions. Table \ref{Table: Human-Network Comparisons} illustrates this and details for one of the images are given in Fig. \ref{Fig: Human-Network Polling Plots} (a). Figure \ref{Fig: Human-Network Polling Plots} (b) shows an image that is consistently misclassified by the network but is almost always correctly classified by humans. This particular example of the digit $7$ is misclassified as a $1$ by several other methods as well \cite{Belongie_2002, Stuhlsatz_2012, Ciresan_2012} and points to the short-comings of current machine learning methods, compared to the human brain. Several images from the $17$ are of poor quality and Fig. \ref{Fig: Human-Network Polling Plots} (c) gives the networks’ predictions and polling results for one such example.

\begin{table}[]
    \centering
    \caption{Comparison of human and network predictions for $6$ out of the $17$ misclassified MNIST test images. For these images, the network predicted labels agree with the human predictions. Polling results are from $120$ individuals who labelled the $17$ misclassified MNIST test images.}
    \begin{tabular}{ccccc}
    \hline
    \textbf{Image Index} & \textbf{GT Label} & \textbf{\begin{tabular}[c]{@{}c@{}}Human/Network Prob.\\ on GT Label\end{tabular}} & \textbf{Network Pred. Label} & \textbf{\begin{tabular}[c]{@{}c@{}}Human/Network Prob.\\ on Pred. Label\end{tabular}} \\
    \hline
    1260 & 7 & 0.38/0.53 & 1 & 0.54/0.64 \\
    1901 & 9 & 0.29/0.18 & 4 & 0.61/0.79 \\
    2597 & 5 & 0.47/0.23 & 3 & 0.51/0.79 \\
    4823 & 9 & 0.42/0.52 & 4 & 0.55/0.71 \\
    5937 & 5 & 0.38/0.55 & 3 & 0.60/0.76 \\
    9729 & 5 & 0.15/0.48 & 6 & 0.82/0.80 \\
    \hline
    \end{tabular}
    \label{Table: Human-Network Comparisons}
\end{table}

\subsection{Other Normalizations}
\label{Section: Other Normalizations}
Max-Min is not unique in its ability to optimally separate the logits. Various other functions can be applied in the routing procedure for CapsNets. We also tested the following functions on the MNIST dataset: 1) Winner-Take-All (WTA), 2) sum, 3) centered Max-Min, 4) Z-score, and 5) adjusted log normalizations. An exhaustive study was not done for each of these methods---our primary goal was to probe the utility of other methods in creating the routing coefficients and whether or not a valid probability distribution was a strict requirement for the assignment of capsules. For WTA, each lower-level capsule in PrimaryCaps only contributes to a single higher-level capsule in DigitCaps. The higher-level capsule assignment is determined by the largest coefficient value for each lower-level capsule. The equation for adjusted log normalization is given by: $c_{ij} = \log(1 + b_{ij} - min(b_{ij}))$ (subtracting the min value and adding one ensures that the min value of the transformed logits is zero). Sum, centered Max-Min, and Z-score normalizations have their usual meanings. A good initialization for each of the five methods is required in order for the network to converge during training. For Max-Min, the initialization for the routing coefficients was robust across two orders of magnitude ($0.01-1.0$). To simplify matters, we initialized routing coefficients to $1.0$ for all five methods listed above.

Table \ref{Table: Normalization Comparisons} shows the test accuracies on the MNIST datasets for the six methods, including Max-Min. Sum normalization performed the worst among the six methods primarily due to difficulties in loss convergence during training---this issue might be alleviated with a more suitable initialization. Centered max-min and adjusted log normalizations performed approximately the same as one another and WTA and Z-Score performed approximately the same. It is worth noting that the WTA method only results in a $0.3\%$ decrease in test accuracy. This is somewhat surprising since WTA assigns a value of $0.0$ for $9$ out of $10$ routing coefficients associated with \textit{each} lower-level capsule. Both centered max-min and Z-score normalization allow routing coefficients to take on negative values. However, the range of values transformed by Z-score is unbounded and can be difficult to initialize properly. The range of values transformed by centered max-min is bounded between $-1$ and $1$. However, it is possible for large negative routing coefficients from centered max-min normalization to counterbalance large positive routing coefficients, leading to a lower network performance compared with Max-Min. Log transformations generally compress high values and spread low values by expressing the values as orders of magnitude and are useful when a high degree of variation exists within variables. This transformation gives decent performance on MNIST but is difficult to initialize properly since the range of the transformed values are not bounded.

\begin{table}[h]
    \centering
    \caption{Test accuracies on the MNIST dataset for several normalization functions. WTA: Winner-Take-All.}
    \begin{tabular}{cc}
    \hline
    \textbf{Normalization} & \textbf{Test Accuracy {[}\%{]}} \\
    \hline
    Max-Min & \textbf{99.55 $\pm$ 0.02} \\
    Centered Max-Min & 99.52 $\pm$ 0.03\\
    Adjusted Log & 99.50 $\pm$ 0.01\\
    WTA & 99.24 $\pm$ 0.36\\
    Z-Score & 99.21 $\pm$ 0.03\\
    Sum & 72.62 $\pm$ 30.05\\
    \hline
    \end{tabular}
    \label{Table: Normalization Comparisons}
\end{table}

\section{Summary}
\label{Section: Summary}
In the formalism from \cite{Sabour_2017}, the logits $b_{ij}$ are converted to the routing coefficients $c_{ij}$ using the Softmax function. For optimal class separation, the routing coefficients should be widely separated in their values. That is, features that are useful for one class, represented by the output of PrimaryCaps, should be strongly coupled to the features in DigitCaps for that class. 

We analyzed the distribution of $c_{ij}$ generated by CapsNets using Softmax and find that they are tightly clustered around the initial value of $0.1$. One reason for this may be that the Softmax function is not scale invariant and for the range of $b_{ij}$’s being produced in the network, Softmax normalization reduces the dynamic range of $c_{ij}$’s. With Max-Min normalization, the dynamic range of the routing coefficients is increased. We demonstrate improved recognition errors, ranging from $8.6$\% to $37.5$\% across five datasets and show that Max-Min allows more routing iterations between adjacent capsule layers without overfitting to the training data. Finally, a single CapsNet is able to achieve a state-of-the-art result on the MNIST test set using just a single model with minimal data augmentation.

\subsubsection*{Acknowledgments}
We would like to thank Peter Dolce for setting up and running the human polling. KPU acknowledges many useful conversations with PS Sastry.

\medskip

\small

\bibliographystyle{plain} 

\appendix
\counterwithin{figure}{section}
\section{Misclassified Images From the MNIST Dataset}
\label{App. Misclassified Images From the MNIST Dataset}
The MNIST images misclassified by each of the three models used in the majority-voting scheme are shown below. Models A and B each misclassified $20$ images out of the $10,000$ test images. Model C misclassified $23$ images. Images with missing pieces of information present the most challenge to the network. Each of the three models are trained using the same set of network parameters and image augmentations mentioned in Section \ref{Section: Performance on MNIST}, with the only difference being the weight initializations for the network layers. The image index, model prediction, and GT labels are listed above each image.

\begin{figure}[h]
\centering
{\includegraphics[width = 5.5 in]{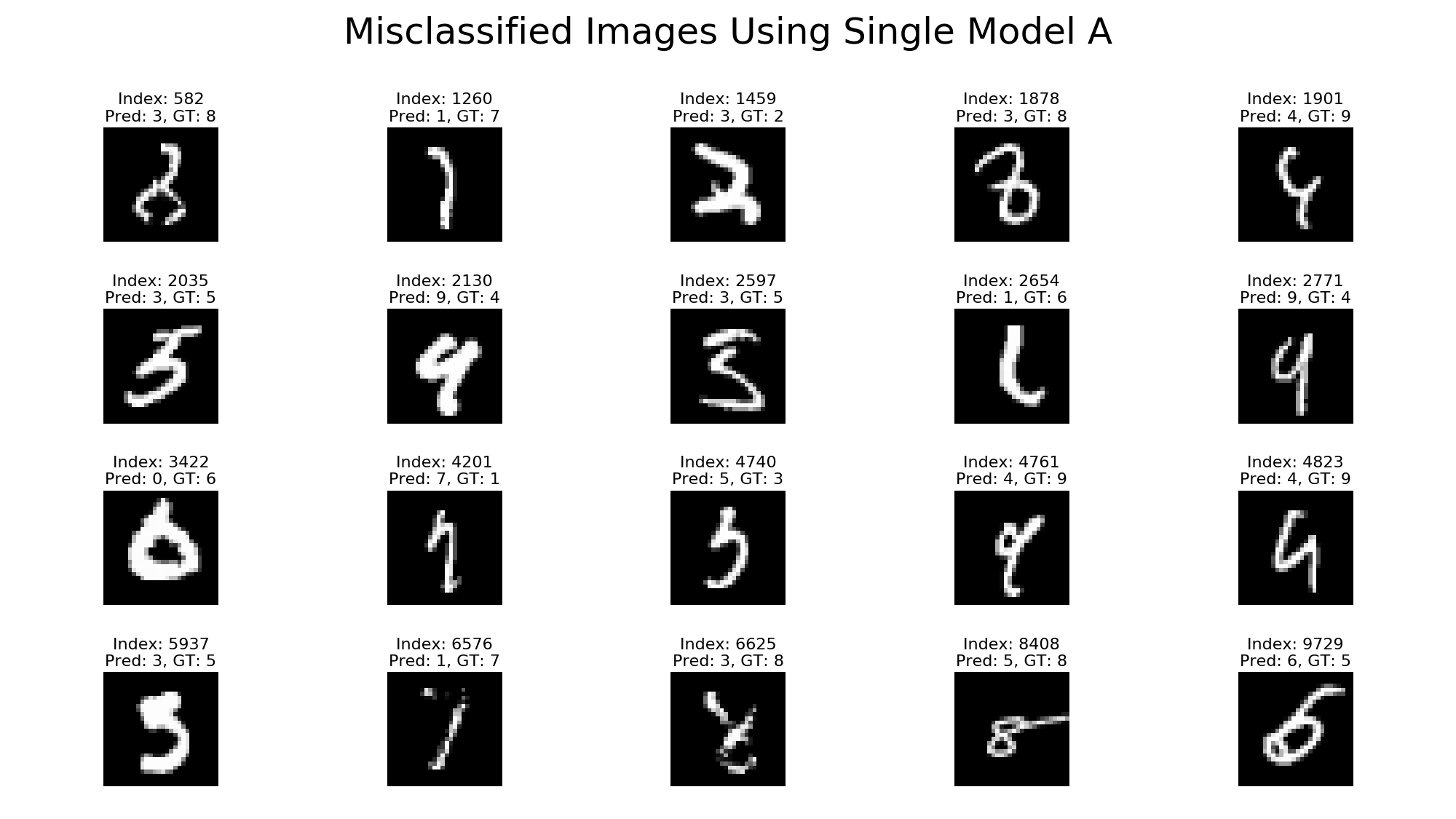}}
\caption{Misclassified MNIST images using CapsNet model A trained using Max-Min normalization.}
\label{Fig: Model A Misclassifications}
\end{figure}

\begin{figure}[h]
\centering
{\includegraphics[width = 5.5 in]{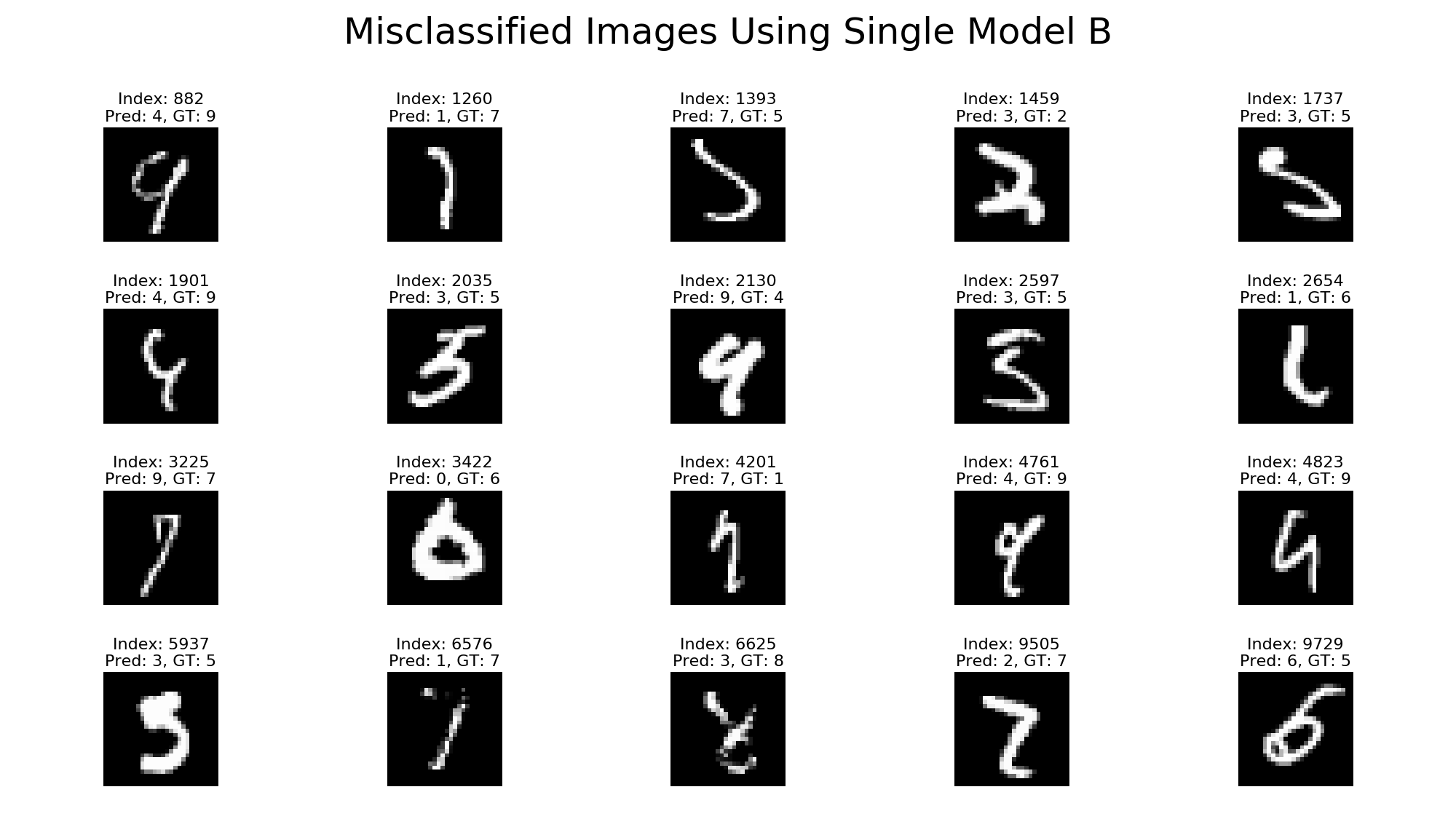}}
\caption{Misclassified MNIST images using CapsNet model B trained using Max-Min normalization.}
\label{Fig: Model B Misclassifications}
\end{figure}

\begin{figure}[h]
\centering
{\includegraphics[width = 5.5 in]{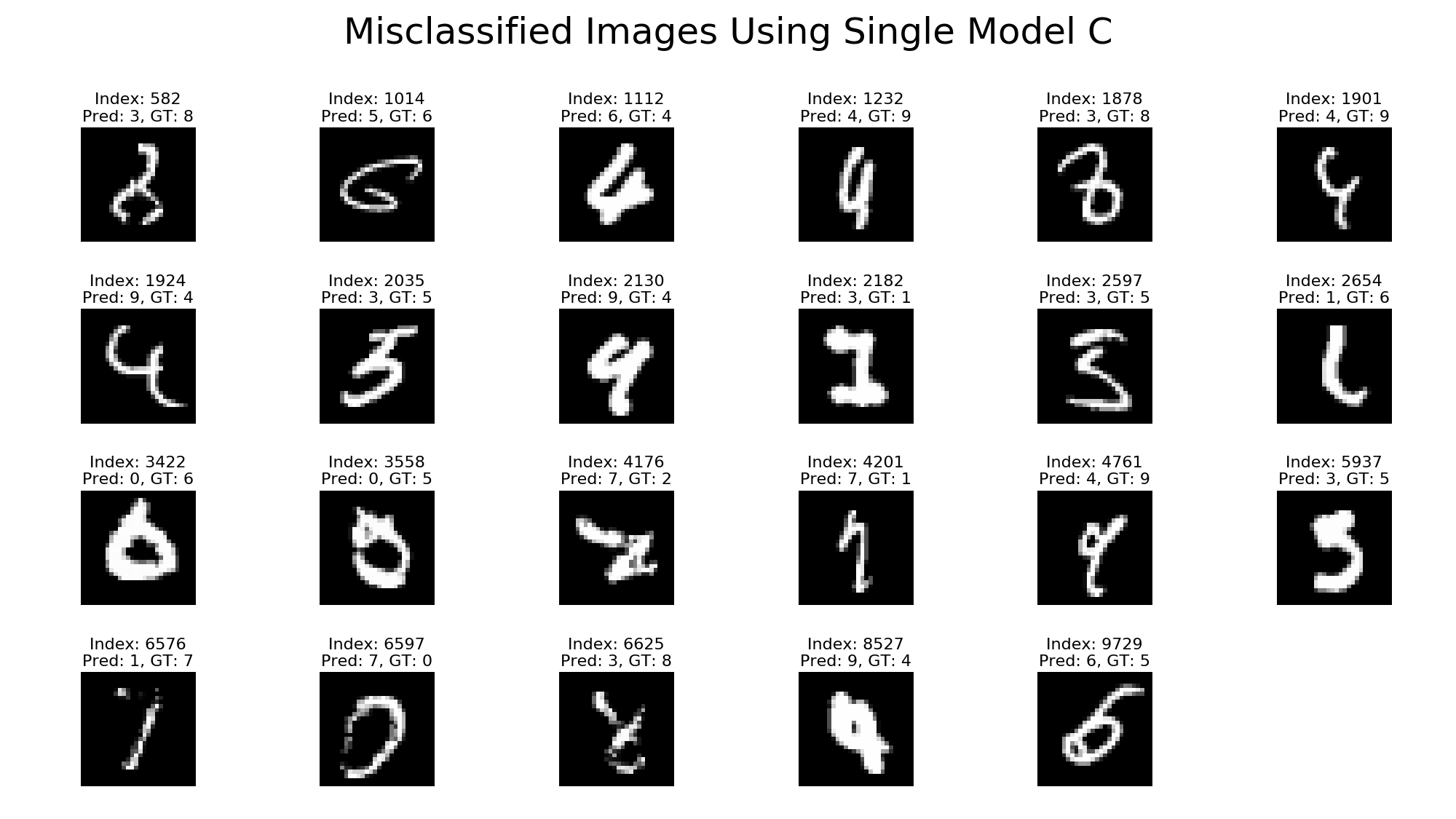}}
\caption{Misclassified MNIST images using CapsNet model C trained using Max-Min normalization.}
\label{Fig: Model C Misclassifications}
\end{figure}

\end{document}